\documentclass{article}

 \usepackage[sglblindworkshop, final]{neurips_2025}
\workshoptitle{Foundations of Reasoning in Language Models}

\usepackage[utf8]{inputenc} 
\usepackage[T1]{fontenc}    
\usepackage{hyperref}       
\usepackage{url}            
\usepackage{amsfonts}       
\usepackage{nicefrac}       
\usepackage{xcolor}         
\usepackage{microtype,amsmath,amssymb,mathtools,booktabs}
\usepackage{algorithm}
\usepackage[noend]{algpseudocode}
\usepackage{slantsc}
\usepackage{autonum}

\title{\textsc{RoBoN}: Routed Online Best-of-n for Test-Time Scaling with Multiple LLMs}

\author{%
  Jonathan Geuter \\
  School of Engineering and Applied Sciences\\
  Harvard University\\
  Kempner Institute for the Study of Natural \& Artificial Intelligence\\
  \texttt{jonathan.geuter@gmx.de} \\
  \And
  Gregor Kornhardt \\
  Department of Mathematics \\
  Technische Universit\"at Berlin \\
  \texttt{g.kornhardt@gmail.com}
}

\begin{document}

\maketitle

\begin{abstract}
Best-of-$n$ is a widely used test-time scaling approach for LLM inference. Yet despite evidence that LLMs exhibit complementary strengths across tasks, traditionally best-of-$n$ relies on a single model to generate responses.
We propose \textsc{RoBoN} (\emph{Routed Online Best-of-$n$}), a \emph{sequential multi-LLM} alternative to the prevailing \emph{single-model} best-of-$n$.
Given a suite of models $\{m_i\}_{i=1}^M$, \textsc{RoBoN} sequentially routes generations one-by-one across models, based on scores computed using a reward model and an agreement signal on the predicted responses. 
This online routing requires no additional training, keeps compute parity, and works with any plug-in reward model.
Across reasoning benchmarks (MATH500, OlympiadBench, MinervaMath, GSM8K, MMLU), \textsc{RoBoN} consistently outperforms standard best-of-$n$ applied to each individual model for larger $n$, with gains of up to 3.4\% in absolute accuracy, and also improves over a uniform multi-model portfolio baseline. Our results indicate that diversity across models can be exploited at inference to improve best-of-$n$ performance over any constituent model alone, providing a simple, training-free path to test-time scaling with multiple LLMs.
\end{abstract}

\section{Introduction}
Large language models (LLMs) have demonstrated remarkable performance across diverse language tasks, and scaling model and data size has long been a reliable way to enhance capabilities~\citep{kaplan2020scalinglawsneurallanguage,geminiteam2024gemini15unlockingmultimodal,openai2024gpt4technicalreport}. However, in recent years there is growing evidence that performance improvements from scaling training compute decelerate~\citep{hernandez2022scalinglawsinterpretabilitylearning,muennighoff2025scalingdataconstrainedlanguagemodels}; furthermore, ever-larger models incur substantial computational and economic costs. This motivated a shift towards \emph{test-time scaling} strategies that spend more inference compute instead of training compute~\citep{snell2024scaling,zhang2025survey,brown2025large}.
Different techniques to scale inference compute exist, such as making models ``think" longer~\citep{muennighoff2025s1,yang2025qwen3technicalreport}, \textit{best-of-$n$} (BoN)~\citep{gao2022scalinglawsrewardmodel,mroueh2024informationtheoreticguaranteespolicy,beirami2025theoreticalguaranteesbestofnalignment}, or \textit{soft BoN}~\citep{verdun2025soft,geuter2025gsi}. 

It is well-known that different LLMs can exhibit complementary strengths~\citep{chen2025harnessingmultiplelargelanguage}, which motives \textit{ensembling} models in some way. Prior work on ensembling multiple LLMs shows that combining models can outperform any individual model~\citep{jiang2023llmblender}, and recent surveys emphasize the opportunity to exploit model diversity at inference~\citep{zhang2025survey}. Yet best-of-$n$ is typically employed with a single model, leaving cross-model BoN strategies unexplored.

\paragraph{Contributions.} We propose \textsc{RoBoN} (\emph{Routed Online Best-of-$n$}), a \emph{sequential, multi-LLM} alternative to single-model BoN. Given a suite $\{m_i\}_{i=1}^M$ of models, and a per-prompt budget $n$, \textsc{RoBoN} routes one generation at a time across models. At each step it evaluates each model’s current head candidate with a scorer that combines a reward model and an agreement signal over predicted answers, commits the best candidate, and recycles the unchosen heads to the next step. This online policy requires no additional training, preserves compute parity (exactly $n$ samples generated in total), and is fully compatible with common acceleration stacks (e.g., vLLM~\citep{vllm}). Across five reasoning benchmarks (MATH500~\citep{lightman2023letsverify}, OlympiadBench~\citep{he2024olympiadbenchchallengingbenchmarkpromoting}, MinervaMath~\citep{lewkowycz2022minerva}, GSM8K~\citep{cobbe2021gsm8k}, MMLU-STEM~\citep{hendrycks2021measuring}), \textsc{RoBoN} consistently improves over BoN with each individual model and over a uniform multi-model portfolio, with gains up to 5\% absolute accuracy.

\section{Background}

\paragraph{Notation.}
Let $\mathcal{V}$ denote a (finite) vocabulary. Let $\mathcal{X}
= \bigcup_{n\in\mathbb{N}} \prod_{i=1}^{n}\mathcal{V}$ be the (countable) space of input prompts, and $\mathcal{Y}
= \bigcup_{n\in\mathbb{N}} \prod_{i=1}^{n}\mathcal{V}$ the (countable) space of responses; while $\mathcal{X}$ and $\mathcal{Y}$ are identical as sets, we distinguish them for clarity. We are given a set of models $\{m_1,...,m_M\}$. For a response generated by model $m_i$, we will typically write a superscript $y^i$ to denote the model that generated the response. If model $m_i$ generates multiple responses, we will distinguish them with subscripts, e.g. $y^i_j$, $j=1,...,n$. Let $\Delta(\mathcal{Y})$ denote the set of probability measures over $\mathcal{Y}$. For $x\in\mathcal{X}$ and model $m$, let $\pi_m(y\mid x)\in\Delta(\mathcal{Y})$ be the model’s conditional distribution. We assume access to a reward model $r:\mathcal{X}\times\mathcal{Y}\to\mathbb{R}$ that assigns a (scalar) outcome score $r(x,y)$ to a response $y$ for prompt $x$. Given a response $y\in\mathcal{Y}$, we extract a normalized answer $a(y)$ (typically, this will appear inside a \texttt{boxed} environment in the answer string $y$) used for agreement statistics; see Section~\ref{sec:robon} for details.
We write $[M]=\{1,\dots,M\}$ for model indices and $n$ for the number of responses per prompt.

\textbf{Test-time scaling and BoN.}
Test-time scaling increases inference-time compute per prompt to improve quality. Broadly, test-time scaling methods can be divided into \textit{parallel} and \textit{sequential} approaches, where the former scales test-time computation in parallel with independently generated outputs, while the latter scales inference via computations that sequentially depend on previous computations.
Various test-time scaling methods have been explored. Sequential approaches include making models 'think' longer by appending \textit{think tokens}~\citep{muennighoff2025s1} or by adding internal thinking to the model outputs~\citep{yang2025qwen3technicalreport}, and self-correction~\citep{qu2024recursive}. Common parallel approaches include majority voting~\citep{wang2023selfconsistencyimproveschainthought}, where $n$ responses are generated in parallel, and the most frequent response is picked; or best-of-$n$~\citep{beirami2025theoreticalguaranteesbestofnalignment}, which instead picks one of the $n$ responses using an auxiliary reward model.
Concretely, given a single model $m$ and reward model $r$, \emph{best-of-$n$} (BoN) draws $y_{1:n}\overset{\text{i.i.d.}}{\sim}\pi_m(\cdot\mid x)$ and outputs 
\[
y^\star \;=\; \arg\max_{j\in[n]} \; r(x,y_j),
\]
which can be seen as a hard-max selection rule. \emph{Soft BoN} instead defines weights
\[
w_j \;\propto\; \exp\{\beta\,r(x,y_j)\},\qquad \beta>0,
\]
and samples $y_j\sim w_j$. As $\beta\to\infty$, soft BoN recovers hard BoN; smaller $\beta$ interpolates toward averaging and can reduce pathologies of extreme selection~\citep{verdun2025soft}. Note that $\beta$ corresponds to an \textit{inverse temperature}. Recent work studies statistical and computational aspects of BoN-style policies, including conditions under which they are (nearly) optimal at fixed inference budgets~\citep{beirami2025theoreticalguaranteesbestofnalignment,huang2025bestofn,geuter2025gsi}.

\textbf{Model ensembling and routing.}
Multiple LLMs often exhibit complementary strengths, motivating \emph{ensembling} and \emph{routing} across models. Output-level ensembling with learned rankers and fusion can outperform any constituent model~\citep{jiang2023llmblender}, and mixture-of-agents (MoA) style approaches aggregate diverse model outputs for additional gains (and debates about when/how to mix)~\citep{jiang2023llmblender,wang2024mixtureofagentsenhanceslargelanguage}.

Routing has been explored at \emph{training time} via Mixture-of-Experts (MoE), where a learned router activates a small subset of experts per token/layer (where \textit{experts} replace the usual feed-forward heads in attention blocks, and typically specialize in certain parts of the domain) to scale performance while controlling active parameter count~\citep{shazeer2017outrageouslylargeneuralnetworks,zhou2022mixtureofexpertsexpertchoicerouting,Cai_2025}.

At \emph{inference time}, \emph{model-level routing} selects among entire LLMs per input, as it is often the case that different LLMs excel at different tasks~\citep{chen2023frugalgptuselargelanguage,shnitzer2023largelanguagemodelrouting,huang2025routerevalcomprehensivebenchmarkrouting}. This usually involves training a parametric router which selects the best model for a given prompt.

Unlike parametric routers that must be trained (either within MoE or as separate model selectors), our approach is \emph{training-free} and operates \emph{online}: we sequentially route a fixed budget of generations across a \emph{portfolio} of off-the-shelf LLMs using only (i) a plug-in reward model and (ii) an agreement signal over predicted answers. This preserves compute parity with single-model BoN (exactly $n$ responses per prompt are generated) while exploiting cross-model diversity without learning a router. However, one key difference to vanilla best-of-$n$ is that our approach falls under the \textit{sequential} test-time scaling umbrella, whereas regular best-of-$n$ is \textit{parallel} in nature.

\section{Routed Online Best-of-$n$}
\label{sec:robon}

\paragraph{Overview.}
Given a portfolio of models $\{m_i\}_{i=1}^M$, a prompt $x$, a per-prompt budget $n$, and a reward model $r(x,y)$, \textsc{RoBoN} allocates the $n$ generations \emph{sequentially}. At each step, it evaluates the \emph{current head} candidate from every model with the reward model (the next unseen sample if that model was chosen last round; otherwise the previously drawn sample is \emph{reused}), computes a marginal score for appending each candidate response-reward pair to the selected set of response-reward pairs $S$, and greedily commits the best one. This “recycle-unchosen-heads” scheduling ensures exact \emph{compute parity}: the procedure draws exactly $n$ responses in total. However, we only iterate $n-M+1$ times, such that the final set of candidate responses only contains $n-M+1$ responses. This corresponds to $n$ generated responses, as in the last iteration, $M-1$ responses are discarded. This is just a technicality to make for a fair comparison to regular best of $n$ (one could use $n+M-1$ in place of $n$ to recover $n$ responses in the final set $S$). Once the final set of response-reward pairs $S$ is generated, the output is chosen by best of $n$ on the set $S$. For details, see Algorithm~\ref{alg:mainalgo}.

\paragraph{Scoring rule.}
Let $S=\{(y_\ell,r_\ell)\}_{\ell=1}^s$ denote the multiset of currently selected candidates with rewards $r_\ell=r(x,y_\ell)$. We define \emph{reward weights} via the temperature-scaled softmax:
\begin{equation}
w_\ell \;=\; \frac{\exp\{\beta r_\ell\}}{\sum_{j=1}^{s}\exp\{\beta r_j\}}
\quad\text{for }\ell=1,\dots,s.
\label{eq:softmax-weights}
\end{equation}
We also extract a normalized answer string $a_\ell \equiv a(y_\ell)$ (e.g., boxed final value for math). Let
\begin{equation}
\mathrm{agree}_S(a_\ell) \;=\; \frac{1}{s}\sum_{j=1}^{s}\mathbf{1}\{a_j=a_\ell\}
\label{eq:agreement}
\end{equation}
denote the (empirical) agreement of $y_\ell$ with the current set $S$. The \emph{agreement-weighted soft score} used to rank candidates is
\begin{equation}
\mathrm{Score}(S;\alpha,\beta)
\;=\;
\sum_{\ell=1}^{s} \, w_\ell \cdot \Big(\alpha\, r_\ell + (1{-}\alpha)\,\mathrm{agree}_S(a_\ell)\Big),
\qquad \alpha\in[0,1].
\label{eq:score}
\end{equation}
At each iteration of Algorithm~\ref{alg:mainalgo}, for each model $i$ we form the tentative set $S\cup\{(y^i,r^i)\}$ by adding its current head and compute the scores
\begin{equation}
\hat{\Delta}_i \;=\; \mathrm{Score}\!\big(S\cup\{(y^i,r^i)\};\alpha,\beta\big).
\label{eq:marginal}
\end{equation}
Intuitively, these scores are a measure of how much value a response $(y^i,r^i)$ adds to the existing set $S$. They can be thought of as an interpolation between best-of-$n$ and majority voting, where the reward term drives selection towards high-reward responses, whereas the agreement term acts as a majority vote incentive, prioritizing responses that occur often in $S$, which can mitigate reward hacking~\citep{skalse2025definingcharacterizingrewardhacking}, as we show in Section \ref{sec:experiments}. A seemingly simpler approach would be to compute $\hat{\Delta}_i=\alpha\, r^i +(1{-}\alpha)\,\mathrm{agree}_{S\cup\{(y^i,r^i)\}}(a(y^i))$ instead. However, surprisingly, we found this to not work well, and the above score to work much better. This might be since it computes a ``marginal" score over the existing set of responses, which estimates the value of \textit{adding} $(y^i,r^i)$ to $S$, instead of the value of $(y^i,r^i)$ by itself.
We note that our algorithm seems to be quite robust to different values of $\alpha$, as long as $\alpha<1$; see Appendix \ref{apx:apx1} for an ablation.

\paragraph{Agreement implementation details.}
Equation~\eqref{eq:agreement} uses normalized answers $a_\ell=a(y_\ell)$, obtained by extracting the solution from the response, and subsequently applying canonical normalizations (such as removing whitespaces and turning everything into lower case). Then, the agreement score $\text{agree}_S(a(y_\ell))$ is computed by comparing the normalized answers for exact matches.

\paragraph{Special cases and limits.}
When $M{=}1$ and $\alpha{=}1$, \textsc{RoBoN} recovers standard soft best-of-$n$ for a single model. For $\alpha{=}0$ the policy becomes agreement-driven, corresponding to majority-style voting. For large $\beta$ and moderate $\alpha$ -- the setting we use in our experiments -- the policy approximates hard best-of-$n$ with an agreement modified reward over a suite of models.

\begin{algorithm}[t]
\caption{\textsc{Routed Online Best-of-$n$ (\textsc{RoBoN})}}
\label{alg:mainalgo}
\begin{algorithmic}[1]
\Require Models $m_1,\ldots,m_M$; prompt $x$; reward model $r$; budget $n$; temperature $\beta>0$, $\alpha>0$.
\vspace{0.2em}
\State $S \gets \emptyset$ \Comment{multiset of selected pairs $(r,a)$}
\State $c_i \gets 1$ for all $i\in[M]$ \Comment{head pointer per model}
\If{$n = 1$} \label{ln:n1}
  \State $i^\star \gets \mathrm{randomChoice}(\{1,...,M\})$
  \State $y^{i^\star}_1\gets m_{i^\star}(\cdot \mid x)$
  \State \textbf{return} $y^{i^\star}_1$
\EndIf
\For{$t=1$ \textbf{to} $n-M+1$} \label{ln:outerT}
  \For{$i=1$ \textbf{to} $M$} \label{ln:inner_i}
    \State $y^i_{c_i}\gets m_i(\cdot\mid x)$ \Comment{generate new response only if $c_i$ increased in previous iteration}
    \State $r^i_{c_i}\gets r(x,y^i_{c_i})$
    \State $\hat{\Delta}_i\gets\textsc{Score}(S\cup \{(y^i_{c_i},r^i_{c_i})\},\alpha,\beta)$
  \EndFor
  \State $i^*\gets\arg\max_i\hat{\Delta}_i$
  \State $S\gets S\cup (y^{i^*}_{c_{i^*}},r^{i^*}_{c_{i^*}})$
  \State $c_{i^*}\gets c_{i^*}+1$ \Comment{increase response counter only for this model}

\EndFor
\State $y\gets \textsc{Best-Of-N}(S)$
\State \textbf{return} $y$
\end{algorithmic}
\end{algorithm}

\begin{algorithm}[t]
\caption{\textsc{Score}$(S,\alpha,\beta)$: agreement-weighted softmax over rewards}
\label{alg:score-agree}
\begin{algorithmic}[1]
\Require Multiset $S=\{(y_\ell,r_\ell)\}_{\ell=1}^s$; weight $\alpha>0$; temperature $\beta>0$.
\State $w \gets \text{softmax}(\beta r)$, \quad $r=(r_1, ..., r_s)$
\For{$\ell=1$ \textbf{to} $s$}
    \State $\text{agree}_S(a(y_\ell))\gets |\{y \in S: a(y_\ell)=a(y)\}| / s$ \Comment{agreement scores on extracted answers $a$}
\EndFor
\State \textbf{return} $\sum_{\ell=1}^{s} w_\ell \cdot (\alpha r_\ell +(1-\alpha)\text{agree}_S(a(y_\ell)))$
\end{algorithmic}
\end{algorithm}

\section{Experiments}\label{sec:experiments}

\textbf{Datasets.} We evaluate on five reasoning datasets: MATH500~\citep{lightman2023letsverify}, OlympiadBench, specifically the \texttt{OE\_TO\_maths\_en\_COMP} split~\citep{he2024olympiadbenchchallengingbenchmarkpromoting}, MinervaMath~\citep{lewkowycz2022minerva}, GSM8K~\cite{cobbe2021gsm8k}, and MMLU-STEM~\citep{hendrycks2021measuring}. While the former are primarily math datasets, we note that MMLU-STEM spans various domains such as biology, physics, computer science, chemistry, and others. For each dataset, we report the average accuracy as well as the 1-sigma confidence intervals.

\textbf{Models ($M{=}4$).} We use the following four models: Qwen2.5-Math-7B-Instruct, DeepSeek-Coder-6.7B-Instruct, Llama-3.1-8B-Instruct, Qwen2.5-Coder-7B-Instruct. We did not perform any search over models, and stuck with our initial suite of models throughout. This suggests \textsc{RoBoN} can work out-of-the-box on suites of models of comparable size, and with potentially even larger benefits for different suites.
We use Skywork/Skywork-Reward-V2-Llama-3.1-8B as the reward model.

\textbf{Implementation.} We implement all models with vLLM~\citep{vllm}. All experiments ran on a single H100 GPU. The implementation is available at \url{https://github.com/j-geuter/RoBoN}, where we also provide the full dataset of all generated responses by all four models on all five datasets, with corresponding rewards and normalized rewards.

\textbf{Hyperparameters.} We set $\alpha=0.4$ (see Appendix \ref{apx:apx1} for an ablation). We found that larger values of $\beta$ work better, so we use $\beta=1e5$ in our experiments, which essentially recovers picking an $\arg\max$. We generate responses with $\texttt{temperature}=1.0$ and $\texttt{top\_p}=0.95$. 

\textbf{Baselines.}
We compare against the following baselines:\\
(a) \textbf{single-model BoN:} We run regular best-of-$n$ on each separate model from the model pool.\\
(b) \textbf{average:} We average the accuracies of the individual best-of-$n$ strategies from (a).\\
(c) \textbf{equal:} We run best-of-$n$, where we assign each model an equal share of $n/M=n/4$ samples (for $n=1$, we pick a model at random).

\begin{figure}
    \centering
    \includegraphics[width=\linewidth]{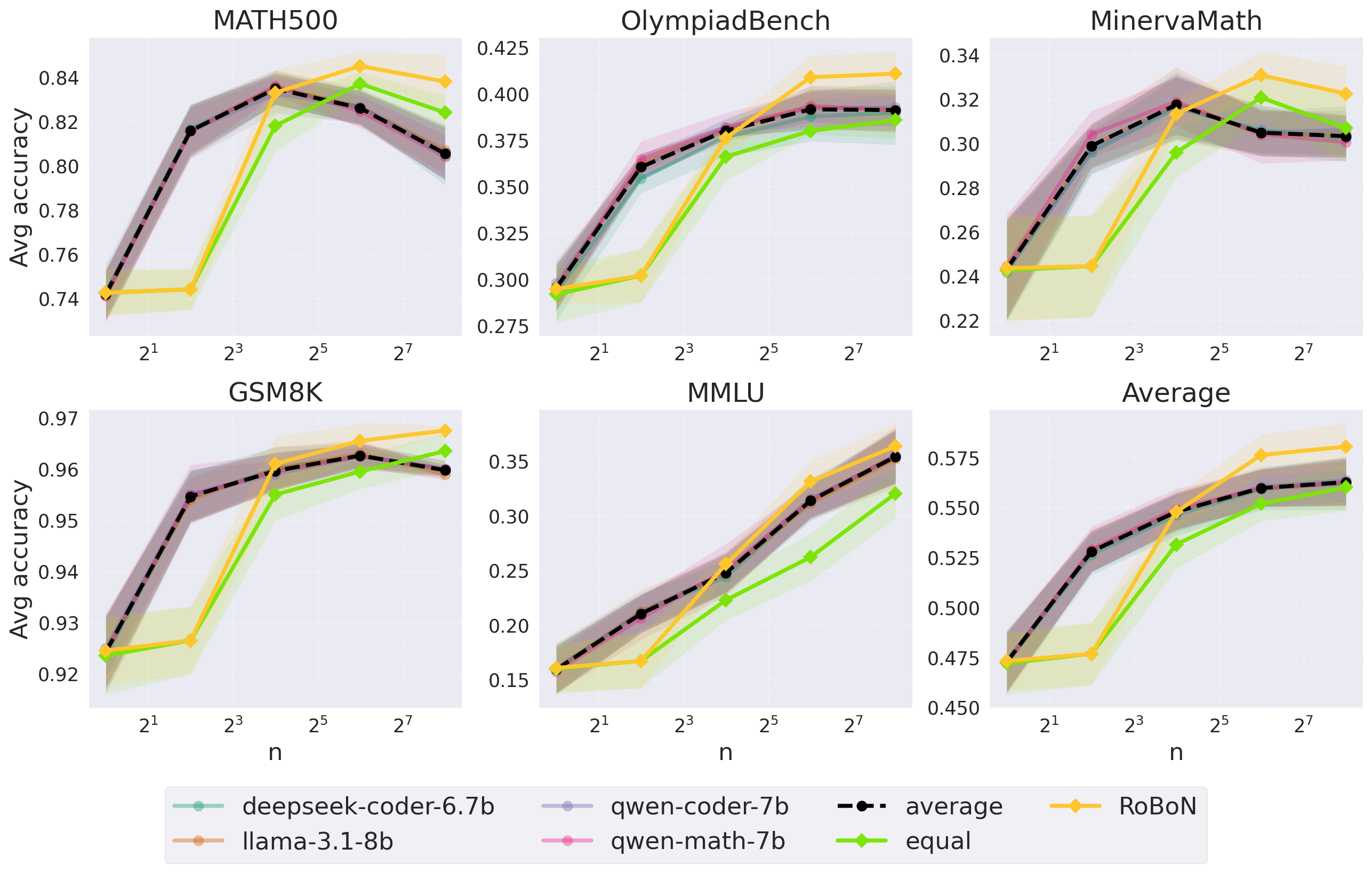}
    \caption{\textbf{\textsc{RoBoN} significantly outperforms best-of-$n$ with individual models for large $n$.} Average accuracies across datasets and methods, with 1-sigma confidence intervals. The degrading performance on some datasets as $n$ increases is likely due to reward hacking~\citep{skalse2025definingcharacterizingrewardhacking}.}
    \label{fig:accuracies}
\end{figure}

\textbf{Reward Normalization.} For each model, we normalize rewards by their empirical CDF. Concretely, starting from a large pre-computed corpus of responses and rewards across datasets, we rank the rewards and map them uniformly onto values in $[0,1]$. This is necessary as otherwise, rewards would not be comparable across models. Indeed, without reward normalization, the performance of \textsc{RoBoN} degrades to that of the average of the individual best-of-$n$ strategies. Note that while this normalization requires a pre-computed dataset of responses and rewards, once this dataset has been created and the empirical CDFs have been estimated, a parametric map from raw to normalized rewards that can be applied to new samples can easily be constructed.

\textbf{Compute, Memory, and Runtime.} In terms of FLOPs, \textsc{RoBoN} is asymptotically exactly en par with standard best-of-$n$ with a single model. The memory requirements grow linearly with the number of models $M$. In terms of runtime, in the worst-case, \textsc{RoBoN} suffers an additional factor of $n$ compared to standard best-of-$n$ with a single model, since \textsc{RoBoN} generates sequentially instead of in parallel. However, this assumes sufficient memory, and if memory is limited, the runtime of best-of-$n$ can be the same as that of \textsc{RoBoN}. Furthermore, as can been seen from Figure \ref{fig:accuracies}, \textsc{RoBoN} achieves accuracies that best-of-$n$ on any single model cannot achieve, no matter how large $n$ is -- in fact, the accuracy of best-of-$n$ often starts deteriorating as $n$ increases beyond a certain point. Hence, even under identical runtime budgets, best-of-$n$ will not achieve the same accuracy as \textsc{RoBoN}. This is the reason we decided not to include an explicit runtime comparison to regular best-of-$n$: In the regime where \textsc{RoBoN} outperforms best-of-$n$, a runtime comparison is not possible, as best-of-$n$ is not able to achieve comparable performance according to our experiments.

\begin{table}[t]
\centering
\small
\caption{Average accuracy across datasets, with 1-sigma confidence intervals.}
\label{tab:avg_datasets}
\setlength{\tabcolsep}{6pt}
\begin{tabular}{lccc}
\toprule
Method & $n=16$ & $n=64$ & $n=256$ \\
\midrule
BoN (deepseek-coder-6.7b) & $0.546 \pm 0.010$ & $0.559 \pm 0.011$ & $0.562 \pm 0.013$ \\
BoN (llama-3.1-8b) & $\mathbf{0.548} \pm 0.010$ & $0.560 \pm 0.009$ & $0.563 \pm 0.012$ \\
BoN (qwen-coder-7b) & $\mathbf{0.548} \pm 0.010$ & $0.560 \pm 0.010$ & $0.564 \pm 0.011$ \\
BoN (qwen-math-7b) & $\mathbf{0.549} \pm 0.010$ & $0.560 \pm 0.009$ & $0.562 \pm 0.011$ \\
Average (across models) & $\mathbf{0.548} \pm 0.009$ & $0.560 \pm 0.009$ & $0.563 \pm 0.012$ \\
equal & $0.532 \pm 0.012$ & $0.552 \pm 0.009$ & $0.560 \pm 0.012$ \\
RoBoN & $\mathbf{0.548} \pm 0.008$ & $\mathbf{0.576} \pm 0.010$ & $\mathbf{0.581} \pm 0.012$ \\
\bottomrule
\end{tabular}
\end{table}

\subsection{Performance of \textsc{RoBoN} on Reasoning Tasks}

In Figure \ref{fig:accuracies}, we report the average accuracies of all baselines and compare them to \textsc{RoBoN}. While \textsc{RoBoN} lags behind baselines for $n=1$ and $n=4$, it is en par for $n=16$ and significantly better at $n=64$ and $n=256$, with accuracy gains of up to $3.4\%$ over baselines. We note best-of-$n$ with individual models performs almost identically across models, yet combining models with \textsc{RoBoN} yields consistent accuracy gains across datasets.\\
The baseline's performance degrades significantly on MATH500 and MinervaMath for larger $n$, likely due to reward hacking. While \textsc{RoBoN} also suffers from reward hacking, the effect is significantly diminished. The results also show that \textsc{RoBoN} achieves accuracies that best-of-$n$ with any single model fails to achieve for \textit{any} $n$. Furthermore, it is interesting to note that \textsc{RoBoN} does not perform well for $n=4$, in which case a single response from one of the four models is selected based on the scores.
Table~\ref{tab:avg_datasets} contains the accuracies for $n=16,64,256$, averaged over datasets. In Appendix~\ref{apx:accuracies}, we provide accuracy results for individual datasets. Appendix \ref{apx:apx1} contains an ablation study over $\alpha$, and Appendix~\ref{apx:shares} details on how often each of the four models is selected by \textsc{RoBoN} in practice.

\section{Discussion and Limitations}
We presented \textsc{RoBoN}, a sequential multi-LLM version of best-of-$n$ sampling for test-time scaling, which adaptively routes generations to models based on scores computed from rewards and agreement signals of responses. In experiments on reasoning datasets, we show that \textsc{RoBoN} significantly outperforms best-of-$n$ baselines of all individual models, as well as a simple portfolio baseline which assigns $n$ samples equally across models. \textsc{RoBoN} is training-free and can be used with any plug-in reward model and across any suite of (comparable) LLMs. However, while sequential test-time scaling methods are often considered advantageous over parallel ones~\citep{muennighoff2025s1}, an obvious drawback of \textsc{RoBoN} over regular best-of-$n$ is runtime cost, if best-of-$n$ is implemented in parallel. Yet, \textsc{RoBoN} achieves accuracies standard best-of-$n$ fails to achieve independent of runtime budget. Furthermore, the nature of the agreement term is such that \textsc{RoBoN} in its current form is only applicable to tasks where it can be immediately verified whether two answers are identical, as is often the case on reasoning datasets. For more open-ended tasks, one could potentially replace this string-based comparison with embedding similarities.\\
In future work, we plan to extend \textsc{RoBoN} to other domains, such as coding, verify its benefits on different suites of models, and develop theoretical guarantees. We note that guarantees for the expected accuracy require further assumptions on the reward model, and even then the nature of the \textsc{RoBoN} scoring algorithm makes it difficult to derive reliable guarantees in practice. Furthermore, semi-parallel versions of \textsc{RoBoN}, that could significantly improve runtime complexity, are an interesting direction for future research. Such variants could e.g. compute part of the responses in parallel, then route the remaining responses to the suite of models based on the already computed responses, and compute them in parallel again.

\section*{Acknowledgements}
 JG is supported by a fellowship from the Kempner Institute for the Study of Natural and Artificial Intelligence at Harvard University.
 
\bibliographystyle{plainnat}
\bibliography{main}

\newpage

\appendix

\section{Appendix}

\subsection{Extended Accuracy Results}\label{apx:accuracies}
In Tables \ref{tab:math500}, \ref{tab:olympiadbench}, \ref{tab:minervamath}, \ref{tab:gsm8k}, and \ref{tab:mmlu}, we report accuracies of \textsc{RoBoN} and all baselines for $n=16, 64, 256$.

\begin{table}[h]
\centering
\small
\caption{\textbf{MATH500.} We report average accuracy with 1-sigma confidence intervals.}
\label{tab:math500}
\setlength{\tabcolsep}{6pt}
\begin{tabular}{lccc}
\toprule
Method & $n=16$ & $n=64$ & $n=256$ \\
\midrule
BoN (deepseek-coder-6.7b) & $0.835 \pm 0.008$ & $0.826 \pm 0.008$ & $0.804 \pm 0.013$ \\
BoN (llama-3.1-8b) & $0.835 \pm 0.008$ & $0.827 \pm 0.008$ & $0.807 \pm 0.011$ \\
BoN (qwen-coder-7b) & $0.832 \pm 0.009$ & $0.826 \pm 0.008$ & $0.806 \pm 0.013$ \\
BoN (qwen-math-7b) & $\mathbf{0.836} \pm 0.004$ & $0.824 \pm 0.007$ & $0.804 \pm 0.010$ \\
Average (across models) & $0.835 \pm 0.007$ & $0.826 \pm 0.008$ & $0.805 \pm 0.012$ \\
equal & $0.818 \pm 0.012$ & $0.837 \pm 0.005$ & $0.824 \pm 0.007$ \\
RoBoN & $0.833 \pm 0.010$ & $\mathbf{0.845} \pm 0.007$ & $\mathbf{0.838} \pm 0.012$ \\
\bottomrule
\end{tabular}
\end{table}

\begin{table}[h]
\centering
\small
\caption{\textbf{OlympiadBench.} We report average accuracy with 1-sigma confidence intervals.}
\label{tab:olympiadbench}
\setlength{\tabcolsep}{6pt}
\begin{tabular}{lccc}
\toprule
Method & $n=16$ & $n=64$ & $n=256$ \\
\midrule
BoN (deepseek-coder-6.7b) & $0.377 \pm 0.010$ & $0.388 \pm 0.013$ & $0.390 \pm 0.017$ \\
BoN (llama-3.1-8b) & $0.380 \pm 0.001$ & $0.393 \pm 0.011$ & $0.391 \pm 0.013$ \\
BoN (qwen-coder-7b) & $0.382 \pm 0.005$ & $0.392 \pm 0.012$ & $0.393 \pm 0.009$ \\
BoN (qwen-math-7b) & $\mathbf{0.382} \pm 0.008$ & $0.393 \pm 0.008$ & $0.391 \pm 0.009$ \\
Average (across models) & $0.380 \pm 0.003$ & $0.392 \pm 0.011$ & $0.391 \pm 0.011$ \\
equal & $0.366 \pm 0.013$ & $0.380 \pm 0.002$ & $0.386 \pm 0.010$ \\
RoBoN & $0.376 \pm 0.007$ & $\mathbf{0.409} \pm 0.012$ & $\mathbf{0.411} \pm 0.012$ \\
\bottomrule
\end{tabular}
\end{table}

\begin{table}[h]
\centering
\small
\caption{\textbf{MinervaMath.} We report average accuracy with 1-sigma confidence intervals.}
\label{tab:minervamath}
\setlength{\tabcolsep}{6pt}
\begin{tabular}{lccc}
\toprule
Method & $n=16$ & $n=64$ & $n=256$ \\
\midrule
BoN (deepseek-coder-6.7b) & $0.316 \pm 0.014$ & $0.306 \pm 0.011$ & $0.302 \pm 0.011$ \\
BoN (llama-3.1-8b) & $0.318 \pm 0.017$ & $0.304 \pm 0.010$ & $0.304 \pm 0.011$ \\
BoN (qwen-coder-7b) & $0.317 \pm 0.015$ & $0.305 \pm 0.009$ & $0.306 \pm 0.011$ \\
BoN (qwen-math-7b) & $\mathbf{0.319} \pm 0.011$ & $0.304 \pm 0.013$ & $0.301 \pm 0.008$ \\
Average (across models) & $0.318 \pm 0.014$ & $0.305 \pm 0.011$ & $0.303 \pm 0.009$ \\
equal & $0.296 \pm 0.011$ & $0.321 \pm 0.011$ & $0.307 \pm 0.015$ \\
RoBoN & $0.313 \pm 0.006$ & $\mathbf{0.331} \pm 0.011$ & $\mathbf{0.323} \pm 0.012$ \\
\bottomrule
\end{tabular}
\end{table}

\begin{table}[h]
\centering
\small
\caption{\textbf{GSM8K.} We report average accuracy with 1-sigma confidence intervals.}
\label{tab:gsm8k}
\setlength{\tabcolsep}{6pt}
\begin{tabular}{lccc}
\toprule
Method & $n=16$ & $n=64$ & $n=256$ \\
\midrule
BoN (deepseek-coder-6.7b) & $\mathbf{0.960} \pm 0.004$ & $0.962 \pm 0.003$ & $0.960 \pm 0.002$ \\
BoN (llama-3.1-8b) & $\mathbf{0.960} \pm 0.004$ & $0.963 \pm 0.003$ & $0.959 \pm 0.001$ \\
BoN (qwen-coder-7b) & $0.959 \pm 0.003$ & $0.962 \pm 0.002$ & $0.960 \pm 0.000$ \\
BoN (qwen-math-7b) & $\mathbf{0.960} \pm 0.003$ & $0.962 \pm 0.003$ & $0.960 \pm 0.002$ \\
Average (across models) & $\mathbf{0.960} \pm 0.004$ & $0.963 \pm 0.002$ & $0.960 \pm 0.001$ \\
equal & $0.955 \pm 0.005$ & $0.960 \pm 0.003$ & $0.964 \pm 0.003$ \\
RoBoN & $\mathbf{0.961} \pm 0.005$ & $\mathbf{0.966} \pm 0.003$ & $\mathbf{0.968} \pm 0.001$ \\
\bottomrule
\end{tabular}
\end{table}

\begin{table}[h]
\centering
\small
\caption{\textbf{MMLU.} We report average accuracy with 1-sigma confidence intervals.}
\label{tab:mmlu}
\setlength{\tabcolsep}{6pt}
\begin{tabular}{lccc}
\toprule
Method & $n=16$ & $n=64$ & $n=256$ \\
\midrule
BoN (deepseek-coder-6.7b) & $0.244 \pm 0.016$ & $0.314 \pm 0.018$ & $0.355 \pm 0.025$ \\
BoN (llama-3.1-8b) & $0.247 \pm 0.019$ & $0.312 \pm 0.015$ & $0.352 \pm 0.027$ \\
BoN (qwen-coder-7b) & $0.249 \pm 0.018$ & $0.316 \pm 0.018$ & $0.354 \pm 0.023$ \\
BoN (qwen-math-7b) & $0.251 \pm 0.024$ & $0.314 \pm 0.015$ & $0.355 \pm 0.026$ \\
Average (across models) & $0.248 \pm 0.017$ & $0.314 \pm 0.016$ & $0.354 \pm 0.025$ \\
equal & $0.223 \pm 0.018$ & $0.262 \pm 0.022$ & $0.320 \pm 0.023$ \\
RoBoN & $\mathbf{0.256} \pm 0.011$ & $\mathbf{0.331} \pm 0.019$ & $\mathbf{0.364} \pm 0.021$ \\
\bottomrule
\end{tabular}
\end{table}

\subsection{Ablation over $\alpha$}\label{apx:apx1}
\begin{figure}[h]
    \centering
    \includegraphics[width=0.6\linewidth]{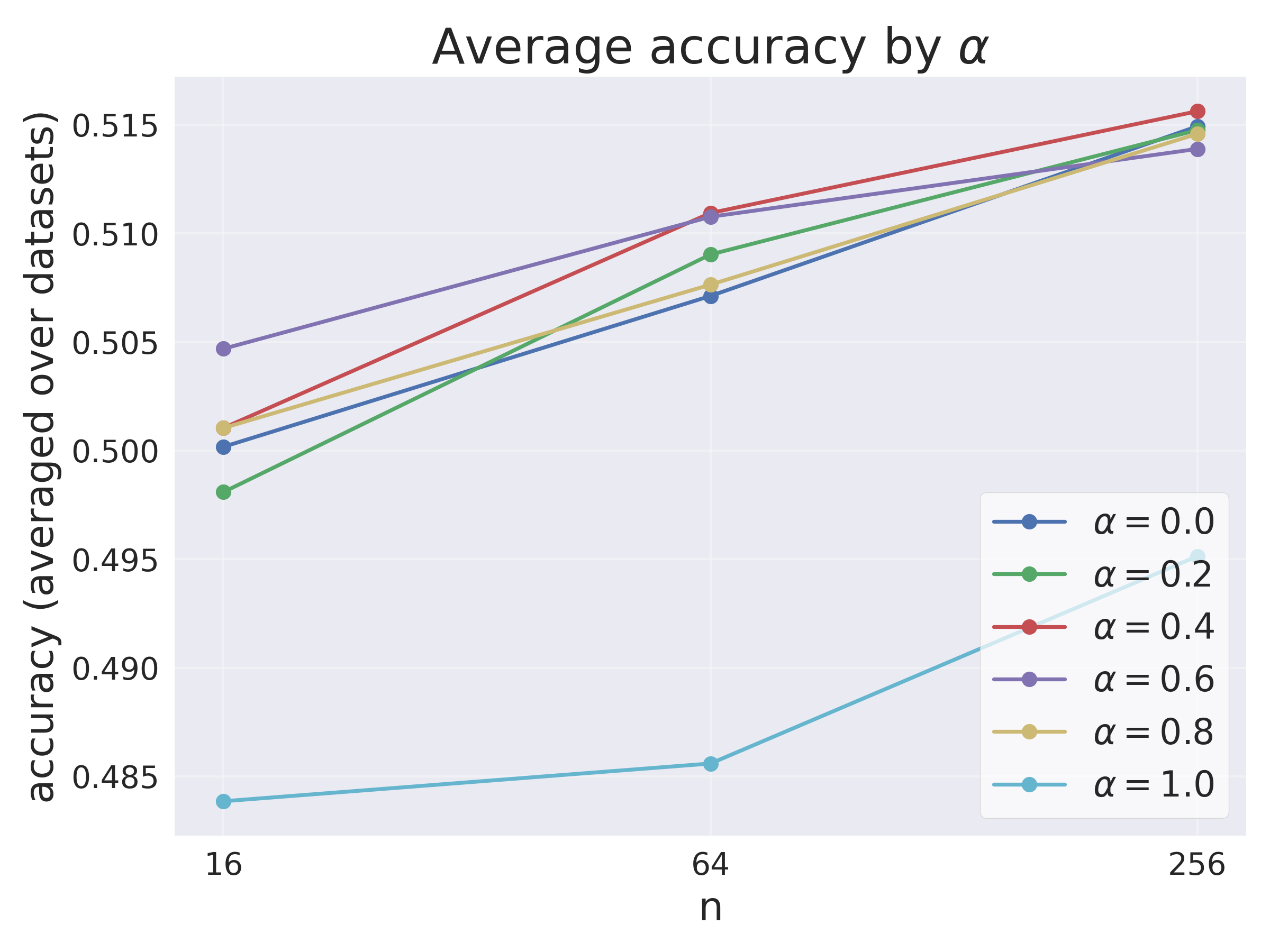}
    \caption{Average accuracy (averaged over MATH500, OlympiadBench, and MinervaMath) for \textsc{RoBoN} with different values of $\alpha$.}
    \label{fig:alphaablation}
\end{figure}

We run an ablation over different values of $\alpha$. In Figure \ref{fig:alphaablation}, we show the average accuracy (over MATH500, OlympiadBench, and MinervaMath) for $\alpha=0.0, 0.2, 0.4, 0.6, 0.8, 1.0$. We can see that most values of $\alpha$ yield similar results, with $\alpha=0.4$ having a slight edge. Only $\alpha=1$, i.e. putting all the weight on the rewards, sees significantly worse performance. This could be attributed to reward hacking~\citep{skalse2025definingcharacterizingrewardhacking}.

\subsection{Model Share across $n$}\label{apx:shares}
\begin{figure}[h]
    \centering
    \includegraphics[width=\linewidth]{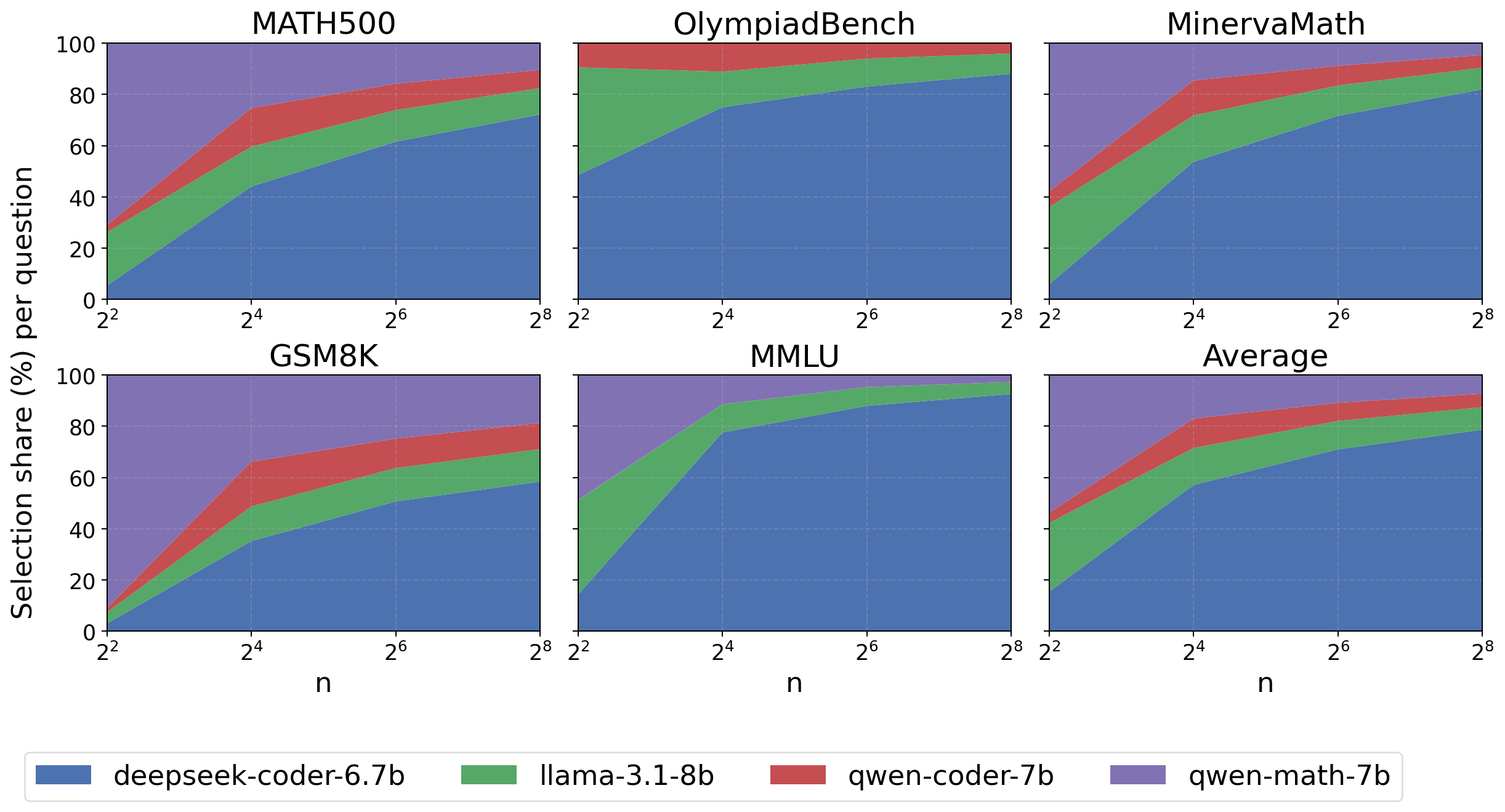}
    \caption{Average share of models selected across different values of $n$ in \textsc{RoBoN}, averaged over datasets. \textsc{RoBoN} selects deepseek-coder-6.7b in the majority of cases; however, \textsc{RoBoN} significantly outperforms this model in terms of accuracy, cmp. Figure \ref{fig:accuracies}.}
    \label{fig:shares}
\end{figure}

In Figure \ref{fig:shares} we show the average share of models selected by \textsc{RoBoN} across datasets and different values of $n$. deepseek-coder-6.7b ranks consistently high with shares between 50-75\% for larger $n$. However, the other three models' share seems to help significantly in boosting performance, as \textsc{RoBoN} performs much better than best-of-$n$ on deepseek-coder-6.7b alone, compare Figure \ref{fig:accuracies}.
We note that \textsc{RoBoN} routes to models differently depending on the dataset: on GSM8K, deepseek-coder-6.7b is selected significantly less often than on average, and on MMLU-STEM, qwen-coder-7b is \textit{never} selected, whereas, interestingly, qwen-math-7b is \textit{never} selected on OlympiadBench. A more thorough investigation of this phenomenon would be an interesting direction for future research.

\clearpage
\section*{NeurIPS Paper Checklist}

\begin{enumerate}

\item {\bf Claims}
    \item[] Question: Do the main claims made in the abstract and introduction accurately reflect the paper's contributions and scope?
    \item[] Answer: \answerYes{} 
    \item[] Justification: A sequential best-of-$n$ method is introduced, as claimed in the abstract.

\item {\bf Limitations}
    \item[] Question: Does the paper discuss the limitations of the work performed by the authors?
    \item[] Answer: \answerYes{} 
    \item[] Justification: We clearly state limitations and future research, such as developing theory, extending the framework to other types of datasets, and mentioning that runtime is affected by the sequential nature of the method (although sequential test-time methods are also considered to have advantages over parallel methods).

\item {\bf Theory assumptions and proofs}
    \item[] Question: For each theoretical result, does the paper provide the full set of assumptions and a complete (and correct) proof?
    \item[] Answer: \answerNA{} 
    \item[] Justification: The paper does not include theoretical results.

    \item {\bf Experimental result reproducibility}
    \item[] Question: Does the paper fully disclose all the information needed to reproduce the main experimental results of the paper to the extent that it affects the main claims and/or conclusions of the paper (regardless of whether the code and data are provided or not)?
    \item[] Answer: \answerYes{} 
    \item[] Justification: All relevant parameters are included, and all source code is provided.

\item {\bf Open access to data and code}
    \item[] Question: Does the paper provide open access to the data and code, with sufficient instructions to faithfully reproduce the main experimental results, as described in supplemental material?
    \item[] Answer: \answerYes{} 
    \item[] Justification: The full source code including a README with instructions is provided.

\item {\bf Experimental setting/details}
    \item[] Question: Does the paper specify all the training and test details (e.g., data splits, hyperparameters, how they were chosen, type of optimizer, etc.) necessary to understand the results?
    \item[] Answer: \answerYes{} 
    \item[] Justification: All details are included.

\item {\bf Experiment statistical significance}
    \item[] Question: Does the paper report error bars suitably and correctly defined or other appropriate information about the statistical significance of the experiments?
    \item[] Answer: \answerYes{} 
    \item[] Justification: We report 1-sigma confidence intervals in figures and tables.

\item {\bf Experiments compute resources}
    \item[] Question: For each experiment, does the paper provide sufficient information on the computer resources (type of compute workers, memory, time of execution) needed to reproduce the experiments?
    \item[] Answer: \answerYes{} 
    \item[] Justification: All compute details are provided.
    
\item {\bf Code of ethics}
    \item[] Question: Does the research conducted in the paper conform, in every respect, with the NeurIPS Code of Ethics \url{https://neurips.cc/public/EthicsGuidelines}?
    \item[] Answer: \answerYes{} 
    \item[] Justification: This paper presents general work in machine learning.

\item {\bf Broader impacts}
    \item[] Question: Does the paper discuss both potential positive societal impacts and negative societal impacts of the work performed?
    \item[] Answer: \answerNA{} 
    \item[] Justification: The paper presents a general test-time scaling algorithm for language models. Language models can be misused; however, we don't feel it is necessary to discuss negative impacts of language models in general in this work.
    
\item {\bf Safeguards}
    \item[] Question: Does the paper describe safeguards that have been put in place for responsible release of data or models that have a high risk for misuse (e.g., pretrained language models, image generators, or scraped datasets)?
    \item[] Answer: \answerNA{} 
    \item[] Justification: There is no high risk of misuse for any of the data we use.

\item {\bf Licenses for existing assets}
    \item[] Question: Are the creators or original owners of assets (e.g., code, data, models), used in the paper, properly credited and are the license and terms of use explicitly mentioned and properly respected?
    \item[] Answer: \answerYes{} 
    \item[] Justification: All datasets we used have been cited. All code is our own.

\item {\bf New assets}
    \item[] Question: Are new assets introduced in the paper well documented and is the documentation provided alongside the assets?
    \item[] Answer: \answerYes{} 
    \item[] Justification: All documentation for our code is included in the code.

\item {\bf Crowdsourcing and research with human subjects}
    \item[] Question: For crowdsourcing experiments and research with human subjects, does the paper include the full text of instructions given to participants and screenshots, if applicable, as well as details about compensation (if any)? 
    \item[] Answer: \answerNA{} 
    \item[] Justification: Not applicable.

\item {\bf Institutional review board (IRB) approvals or equivalent for research with human subjects}
    \item[] Question: Does the paper describe potential risks incurred by study participants, whether such risks were disclosed to the subjects, and whether Institutional Review Board (IRB) approvals (or an equivalent approval/review based on the requirements of your country or institution) were obtained?
    \item[] Answer: \answerNA{} 
    \item[] Justification: Not applicable.

\item {\bf Declaration of LLM usage}
    \item[] Question: Does the paper describe the usage of LLMs if it is an important, original, or non-standard component of the core methods in this research? Note that if the LLM is used only for writing, editing, or formatting purposes and does not impact the core methodology, scientific rigorousness, or originality of the research, declaration is not required.
    \item[] Answer: \answerNA{} 
    \item[] Justification: LLMs have not been used in such ways for this paper.

\end{enumerate}

\end{document}